\UseRawInputEncoding
\documentclass{article}
\usepackage{spconf,amsmath,graphicx}
\usepackage{amssymb}
\usepackage{epstopdf}
\usepackage{cite}
\usepackage{overpic}
\usepackage{bm}
\usepackage{multirow}
\usepackage{color}
\usepackage{multirow}
\usepackage{algorithm}
\usepackage{algorithmic}
\usepackage{multicol}
\usepackage{siunitx}
\usepackage{amsthm}
\usepackage{makecell}
\usepackage{bbding}
\usepackage{balance}

\title{Self-supervised Contrastive Learning for Audio-Visual Action Recognition}
\name{Yang Liu$^{1}$  \qquad   Ying Tan$^{1}$  \qquad  Haoyuan Lan$^{1}$}
\address{$^{1}$ Sun Yat-sen University, China}
\begin{document}
%
\maketitle
\begin{abstract}
The underlying correlation between audio and visual modalities can be utilized to learn supervised information for unlabeled videos. In this paper, we propose an end-to-end self-supervised framework named Audio-Visual Contrastive Learning (AVCL), to learn discriminative audio-visual representations for action recognition. Specifically, we design an attention based multi-modal fusion module (AMFM) to fuse audio and visual modalities. To align heterogeneous audio-visual modalities, we construct a novel co-correlation guided representation alignment module (CGRA). To learn supervised information from unlabeled videos, we propose a novel self-supervised contrastive learning module (SelfCL). Furthermore, we build a new audio-visual action recognition dataset named Kinetics-Sounds100. Experimental results on Kinetics-Sounds32 and Kinetics-Sounds100 datasets demonstrate the superiority of our AVCL over the state-of-the-art methods on large-scale action recognition benchmark.
\end{abstract}
\section{Introduction}
\label{sec:intro}

With the development of deep learning, many visual recognition tasks \cite{zhu2022hybrid,liu2022causal,liu2022cross,TFP,VLCI} have achieved state-of-the-art performance. This can be primarily attributed to the learned rich representation from well-trained networks using large-scale image/video datasets with strong supervision information \cite{7,9}. However, annotating such large-scale data is laborious, expensive, and impractical, especially for data-driven high-level tasks like video action recognition \cite{liu2018transferable,liu2018global,liu2018hierarchically,liu2019deep,liu2021semantics}. To fully leverage the large amount of unlabeled data, self-supervised learning  gives a reasonable way to utilize unlabeled data to obtain supervision signals.

Self-supervised learning generates supervision signals by adopting various pretext tasks or contrastive learning methods \cite{liu2021temporal}. The pretext task learning methods for the video data includes Sorting sequences \cite{12}, Order prediction \cite{14}, and Temporal Contrastive Graph Learning \cite{TCGL}, etc. Besides, many contrastive learning methods have been proposed, such as the NCE \cite{19}, MoCo \cite{20}, SimCLR \cite{22}, SimSam \cite{23}. However, the existing self-supervised learning methods usually focus on single modality and ignore heterogeneous alignment and fusion of multiple modalities.

Actually, the audio is an important modality accompanied by the visual content of the video \cite{DBLP}, which contains complementary information that can alleviate the problem that some video samples are indistinguishable in terms of appearance (e.g., playing the guitar and violin) \cite{Epic-fusion}. To fully leverage the knowledge from audio, the audio-visual representation learning \cite{Listen_to_look,AVC,L3,16,Look_listen_and_attend,AVS} has attracted increasing attention recently. However, most of these methods focus on labelled audio-visual data and ignore the large-scale unlabelled data. Since the audio-visual co-occurrences provide free supervised signals, some methods utilized this attribute to co-train a joint model and learn cross-modal representations. For instance, AVE-Net \cite{16} trained a audio-visual model to find the visual area with the maximum similarity for the current audio clip. CMA \cite{Look_listen_and_attend} proposed a co-attention framework to learn generic cross-modal representations from unlabelled videos. However, previous methods are pervasively limited by the huge heterogeneous modality differences of audio-visual modalities.

In this paper, we focus on how to efficiently learn discriminative self-supervised audio-visual representations. Specifically, we consider audio as the primal modality to provide complementary information for visual modalities (RGB images) in action recognition and propose a novel contrastive learning algorithm framework named Audio-Visual Contrastive Learning (AVCL) to learn cross-modality features with strong representation ability. The AVCL is a self-supervised framework that integrates multi-modality alignment, multi-modal fusion and contrastive learning into an end-to-end manner. The main contributions of this paper are as follows:
1) To fully fuse complementary information between audio and visual modalities, we design an attention based multi-modal fusion module (AMFM). 2) To align heterogeneous audio-visual data, we construct a co-correlation guided representation alignment module (CGRA). 3) To learn discriminative unsupervised audio-visual representations, we propose a self-supervised contrastive learning module (SelfCL). 4) We build a new dataset named Kinetics-Sounds100 to expand the existing audio-visual datasets.

\begin{figure*}
\centerline{\includegraphics[scale=0.5]{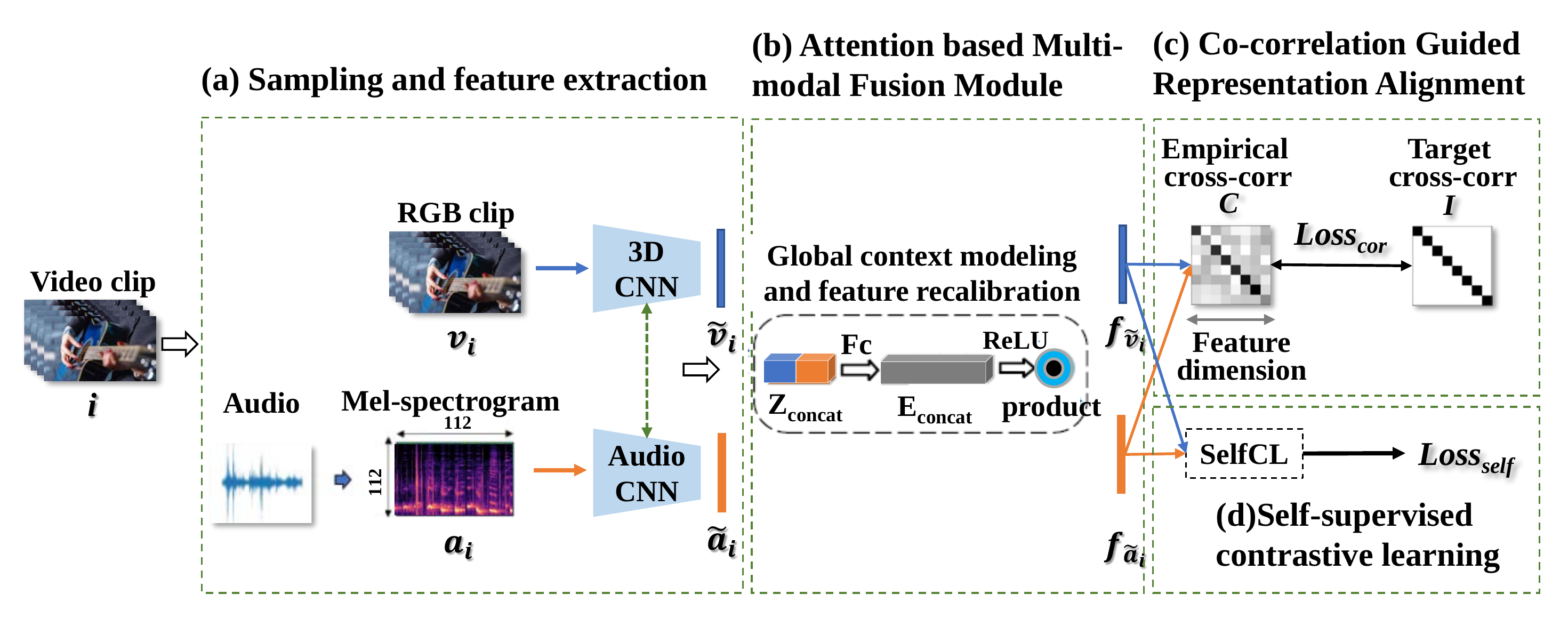}}
\vspace{-10pt}
\caption{The framework of the  Audio-Visual Contrastive Learning (AVCL): (a) Sampling and Feature Extraction; (b) Attention based Multi-modal Fusion Module (AMFM); (c) Co-correlation Guided Representation Alignment (CGRA); (d) Self-supervised Contrastive Learning (SelfCL).}
\vspace{-10pt}
\end{figure*}

\section{METHODOLOGY}
\label{sec:format}
Given a video, the clips from this video are composed of frames with the size $c\times l\times h\times w$, where $c$ is the number of channels, $l$ is the number of frames, $h$ and $w$ indicate the height and width of frames. The size of the 3D convolutional kernel is $t\times d\times d$, where $t$ is the temporal length and $d$ is the spatial size. We define the sequence of video RGB tuples as $V=\left<v_1,v_2,...,v_i,...,v_N\right>$, where $v_i$ is the RGB video clip generated by continuously sampling $m$ frames from a video. The audio mel-spectrogram tuples is represented as $A=\left<a_1,a_2,...,a_i,...,a_N\right>$, where $a_i$ is a mel-spectrogram generated from audio extracted from the video $v_i$, and $\tilde{y}_i$ is the category label of the video $v_i$.

\vspace{-10pt}
\subsection{Sampling and Feature Extraction}
{\textbf{RGB Processing}:} We randomly sample consecutive frames from the video to construct video RGB clips $v_i \left(i=1,...,N\right)$. For RGB clips, they contain dynamic information and strict temporal dependency of the videos.

{\textbf{Audio Processing}:} Different from the  previous work that only extract 1.28s of audio, we extract the whole audio and convert it to single-channel, and resample it to 24kHz. Then we convert it to a mel-spectrogram representation using an STFT of window length 10ms, hop length 10ms and 256 frequency bands. This results in a 2D spectrogram $a_i$ matrix of size 256$\times$256, after which we scale it by the mel-scale.

{\textbf{Feature Extraction}:} Although the audio modality has converted to a mel-spectrogram, it is still quite different from the RGB image due to the huge heterogeneous modality differences between the audio and RGB image representation. Therefore, using only one model to jointly learn the audio-visual is not feasible. To address this issue, we use two independent models with the same basic framework to extract the features for these two modalities, respectively. The parameters between these two models are not shared. The extracted feature for RGB clip $v_i$ is denoted as $\tilde{v}_i$, the corresponding feature for audio mel-spectrogram $a_i$ is denoted as $\tilde{a}_i$.

\vspace{-10pt}
\subsection{Attention based Multi-modal Fusion Module}
To better fuse audio and visual modalities, we propose an Attention based Multi-modal Fusion Module (AMFM). Since the features from different modalities are correlated, we construct a cross-modal feature fusion module that receives features from different modalities and learns a global context embedding, then this embedding is used to recalibrate the input features from different modalities.

To utilize the correlation between these two modalities, we concatenate these two feature vectors and get joint representations through a fully-connected layer:
\begin{equation}
    Z_u = W_s\left[\tilde{v}_i, \tilde{a}_i\right]+b_s
    \label{eq:3-1}
\end{equation}
where $\left[\cdot,\cdot\right]$ denotes the concatenation operation, $Z_u\in\mathbb{R}^{c_u}$ denotes the joint representation, $c_u$ is defined as the half of the sum of dimensionality of $\tilde{v}_i$ and $\tilde{a}_i$, to restrict the model capacity and increase its generalization ability. $W_s$ and $b_s$ are weights and bias of the fully-connected layer. To make use of the global context information aggregated in the joint representations $Z_u$, we predict excitation signal via a fully-connected layer:
\begin{equation}
    E = W_e Z_u + b_e
    \label{eq:3-2}
\end{equation}
where $W_e$ and $b_e$ are weights and biases of the fully-connected layer. After obtaining the excitation signal $E\in\mathbb{R}^{c}$, we use it to recalibrate the input feature $\tilde{v}_i$ and $\tilde{a}_i$ adaptively by a simple gating mechanism:
\begin{equation}
    f_{\tilde{v}_i}=\sigma\left(E\right)\odot\tilde{v}_i
    \label{eq:3-3}
\end{equation}
\begin{equation}
    f_{\tilde{a}_i}=\sigma\left(E\right)\odot\tilde{a}_i
    \label{eq:3-4}
\end{equation}
where $\odot$ is channel-wise product operation for each element in the channel dimension, and $\sigma\left(\cdot\right)$ is the ReLU function. $W_c$ and $b_c$ are weights and biases of the fully-connected layer. In this way, we can allow the features of one modality to recalibrate the features of another modality while concurrently preserving the correlation among different modalities.

\vspace{-10pt}
\subsection{Co-correlation Guided Representation Alignment}
To align heterogeneous audio-visual multi-modality data, we introduce a novel Co-correlation Guided Representation Alignment module (CGRA), which aims to strengthen the correlation between two modalities by making the cross-correlation matrix between the network outputs as close to the identity matrix as possible.
\begin{equation}
Loss_{cor} \triangleq \sum_i\left(1-C_{ii}\right)^2+ \lambda{\sum_i\sum_{j\neq i}C_{ij}^2}
  \label{eq:3-5}
\end{equation}
where $\lambda$ is a positive constant trading off the importance of the first and second terms of the loss, and where $C$ is the cross-correlation matrix computed between the outputs of the two identical networks along the batch dimension:
\begin{equation}
   C_{ij} \triangleq \frac{\sum_bf_{\tilde{v}_i^b}f_{\tilde{a}_j^b}}{\sqrt{\sum_b\left(f_{\tilde{v}_i^b}\right)^2}\sqrt{\sum_b\left(f_{\tilde{a}_j^b}\right)^2}}
  \label{eq:3-6}
\end{equation}
where b denotes batch samples. $C$ is a square matrix with the same size as the network¡¯s output, and with values between -1 (perfect anti-correlation) and 1 (perfect correlation).

\vspace{-10pt}
\subsection{Self-supervised Contrastive Learning}
To learn supervised information from unlabeled videos, we propose a novel self-supervised contrastive learning module (SelfCL). The core of SelfCL is to make the distance between features from the same sample as close as possible, and the features from different samples as far away as possible. Here, we take the RGB clip feature $v_i$ of the video $O_i$ and the mel-spectrogram $a_i$ generated by the corresponding audio as examples to clarify our SelfCL. In RGB modality, the contrastive loss is defined as:
\begin{equation}
\begin{gathered}
    Loss_{v}^{self} = \sum_{i \in I} Loss_{f_{v_i}}^{self} \\
    = -\sum_{i \in I} \log \frac{h_\theta \left(f_{\tilde{v}_i} \cdot f_{\tilde{a}_i}\right)}{\sum_{j \in I} h_\theta \left(f_{\tilde{v}_i} \cdot f_{\tilde{a}_j} \right)+\sum_{j \in A\left(i\right)} h_\theta \left(f_{\tilde{v}_i} \cdot f_{\tilde{v}_j}\right)}
    \label{eq:3-7}
\end{gathered}
\end{equation}

Here, $i\in I\equiv\left\{1,...,N\right\}$, $A\left(i\right)\equiv I\setminus \left\{i\right\}$, $\tau\in R^+$ is a scalar temperature parameter. $h_\theta$ is the distance between two features, which is obtained by using the feature representation of non-parametric softmax:
\begin{equation}
    h_\theta \left(f_{\tilde{v}_i} \cdot f_{\tilde{a}_j}\right)=exp\left(\frac{f_{\tilde{v}_i}\cdot f_{\tilde{a}_j}}{\parallel f_{\tilde{v}_i}\parallel \cdot \parallel f_{\tilde{a}_j} \parallel}\cdot\frac{1}{\tau}\right)
    \label{eq:3-8}
\end{equation}

Similarly, the contrastive loss for audio modality can be computed as:
\begin{equation}
    \begin{gathered}
    Loss_{a}^{self} =\sum_{i \in I} Loss_{f_{a_i}}^{self} \\
    = -\sum_{i \in I} \log \frac{h_\theta \left(f_{\tilde{a}_i} \cdot f_{\tilde{v}_i}\right)}{\sum_{j \in I} h_\theta \left(f_{\tilde{a}_i} \cdot f_{\tilde{v}_j} \right)+\sum_{j \in A\left(i\right)} h_\theta \left(f_{\tilde{a}_i} \cdot f_{\tilde{a}_j}\right)}
    \end{gathered}
    \label{eq:3-9}
\end{equation}

Then, the overall contrastive loss is defined
as follows:
\begin{equation}
    Loss_{self} = Loss_{v}^{self} + Loss_{a}^{self}
    \label{eq:3-10}
\end{equation}

The overall loss for our Audio-Visual
Contrastive Learning (AVCL) framework is obtained by combing Eq. (\ref{eq:3-5}) and Eq. (\ref{eq:3-10}), where $\lambda_{cor}$ and $\lambda_{self}$ control the contribution of $Loss_{cor}$ and $Loss_{self}$, respectively:
\begin{equation}
    Loss = \lambda_{cor}Loss_{cor} + \lambda_{self}Loss_{self}
    \label{eq:3-13}
\end{equation}

\vspace{-20pt}
\section{EXPERIMENTS}

\vspace{-10pt}
\subsection{Datasets}
We conduct experiments on two video datasets, namely, Kinetics-Sounds32 \cite{L3} and Kinetics-Sounds100.

\textbf{Kinetics-Sounds32 (K32).} This dataset contains 19.2k videos (16.1k training, 1.1k validation, 2.0k test) from the Kinetics dataset with 34 action classes, which have been chosen to be potentially manifested visually and aurally.

\textbf{Kinetics-Sounds100 (K100).} To expand the existing audio-visual action recognition datasets and better evaluate our algorithm, we build a new audio-visual action recognition dataset named Kinetics-Sounds100. We take a subset from the Kinetics400 dataset \cite{29}, which contains YouTube videos manually annotated for human actions using Mechanical Turk, and cropped to 10 seconds around the action. The subset contains 61.3k videos (50.2k training, 3.9k validation, 7.2k testing) from the Kinetics400 dataset with 100 action classes. Including all the categories from K32, we add some challenging categories with similar appearance but different audio features,  to increase the generalizability of the model. For example, dance (dancing charleston and belly dancing), sports (playing volleyball and playing cricket), and jubilation (applauding and celebrating), etc.

\vspace{-10pt}
\subsection{Experimental Setting}
\textbf{Network Architecture.} For video encoder, C3D\cite{c3d}, R3D\cite{r3d} and R(2+1)D\cite{r21d} are used as backbones, where the kernel size of 3D convolutional layers is set to $3\times 3 \times 3$. The R3D network is implemented with no repetitions in conv{2-5}, which results in 9 convolution layers totally. The C3D network is modified by replacing the two fully connected layers with global spatiotemporal pooling layers. The R(2+1)D network has the same architecture as the R3D with only 3D kernels decomposed. Dropout layers are applied between fully-connected layers with p = 0.7. The feature extraction network output of the pre-training process is 512 channels.

\textbf{Parameters.} Following the settings in \cite{14}, \cite{30}, we set the clip length as 16. RGB images and mel-spectrograms are randomly cropped to 112$\times$112. Two independent models with the same backbone are utilized for two modalities, respectively. We set the parameters $\lambda=0.005$ for $Loss_{cor}$, and $\lambda_{cor}=0.9$, $\lambda_{self}=0.1$ to balance the contribution between CGRA module and SelfCL module. We use mini-bach SGD with the batchsize 16, the initial learning rate 0.001, the momentum 0.9 and the weight decay 0.0005 in training process, while the weight decay is set to 0.005 in finetune and test process. The training process lasts for 300 epochs, while the finetune and test lasts for 150 epochs. The model with the best validation accuracy is saved to the best model.

\vspace{-10pt}
\subsection{Ablation Study}
In this section, we conduct ablation studies with R3D as the backbone, to analyze the contribution of each component of our AVCL and some important hyperparameters.

\textbf{The sampling methods of audio-visual features.} From Table \ref{tab:4-2}, we can see that our sampling method performs better than other methods, which verifies that the whole audio contains more comprehensive information and the RGB clip contains dynamic information and strict temporal relation.

\begin{table}[!t]
  \centering
  \caption{Action recognition accuracy $\left(\%\right)$ with different sampling methods and different number of models on K32.}
  \small
  \setlength{\tabcolsep}{2pt}
  \centering
  \begin{tabular}{cccc}
    \hline
    RGB length & Audio length & Model number & Accuracy    \\
     \hline
    16 & Raw audio & 2 & 50.2     \\
    1  & 1.2s Mel-spectrogram & 2 & 67.9     \\
    16  & 1.2s Mel-spectrogram & 2 & 72.1     \\
    16 & \textbf{Whole Mel-spectrogram} & \textbf{2} & \textbf{75.6}  \\
    16 & Whole Mel-spectrogram & 1 & 60.5  \\
    \hline
  \end{tabular}
  \vspace{-20pt}
    \label{tab:4-2}
\end{table}

\begin{table}[t]
  \centering
  \caption{Action recognition accuracy $\left(\%\right)$ with/without different layer of CGRA module on K32.}
  \centering
  \small
  \setlength{\tabcolsep}{3pt}
  \begin{tabular}{cccc}
    \hline
    Backbone & CGRA & Layer & Accuracy    \\
    \hline
    R3D & \XSolidBrush & None  & \textbf{72.5}    \\
    R3D & \checkmark & conv3 & 68.5    \\
    R3D & \checkmark & conv4 & 69.3    \\
    R3D & \checkmark & conv5 & 70.9    \\
    R3D & \checkmark & \textbf{pool-final}&\textbf{75.6}   \\
    R3D & \checkmark & conv5+pool & 70.4    \\
    \hline
  \end{tabular}
    \vspace{-10pt}
    \label{tab:4-4}
\end{table}

\begin{table}[t]
  \centering
  \caption{Action recognition accuracy $\left(\%\right)$ with/without self-supervised contrastive learning module, with different values of $\lambda_{cor}$ and $\lambda_{self}$, and with/without AMFM module on K32.}
  \centering
  \small
  \setlength{\tabcolsep}{3pt}
  \begin{tabular}{cccccc}
    \hline
    Backbone & SelfCL & $\lambda_{cor}$ & $\lambda_{self}$ & AMFM & Accuracy    \\
    \hline
    R3D & \XSolidBrush & 1   & 0  & \checkmark & \textbf{73.5}     \\
    R3D & \checkmark & 0.1 & 0.9 & \checkmark & 72.0    \\
    R3D & \checkmark & 0.3 & 0.7 & \checkmark & 72.2     \\
    R3D & \checkmark & 0.5 & 0.5 & \checkmark & 74.1     \\
    R3D & \checkmark & 0.7 & 0.3 & \checkmark & 73.7     \\
    R3D & \checkmark & \textbf{0.9} & \textbf{0.1} & \checkmark & \textbf{75.6}  \\
    R3D & \checkmark & \textbf{0.9} & \textbf{0.1} & \XSolidBrush & \textbf{73.5}  \\
    \hline
  \end{tabular}
    \vspace{-20pt}
    \label{tab:4-5}
\end{table}

\begin{table}[!t]
  \centering
  \caption{Comparison with the state-of-the-art self-supervised contrastive learning methods on K32 and K100.}
  \small
  \setlength{\tabcolsep}{1pt}
  \begin{tabular}{cccccc}
  \hline
  \centering
    Method  & Pretrain & Test & Backbone & K32  & K100   \\
    \hline
    Baseline-R  & RGB & RGB & R3D & 60.4 & 53.6     \\
    Baseline-A  & Audio & Audio & R3D & 52.9 & 40.9     \\
    \textbf{Baseline-R+A} & RGB+Audio & RGB+Audio & R3D & \textbf{73.9} & \textbf{61.3}     \\
    IIC \cite{28} & RGB+Audio & RGB+Audio & R3D & 60.5 & 46.4     \\
    SimCLR \cite{22} & RGB+Audio & RGB+Audio & R3D & 69.7 & 59.1     \\
    MoCo \cite{20} & RGB+Audio & RGB+Audio & R3D & 68.2 & 58.2     \\
    \textbf{BYOL} \cite{21} & RGB+Audio & RGB+Audio & R3D & \textbf{73.3} & \textbf{61.4}     \\
    SimSam \cite{23} & RGB+Audio & RGB+Audio & R3D & 72.9 & 60.8     \\
    Barlow Twins \cite{24} & RGB+Audio & RGB+Audio & R3D & 69.6 & 59.3     \\\hline
    \textbf{AVCL(Ours)} & RGB+Audio & RGB+Audio & R3D & \textbf{75.6} &   \textbf{64.4}   \\
    \textbf{AVCL(Ours)} & RGB+Audio & RGB+Audio & C3D & \textbf{74.0} &   \textbf{61.3} \\
    \textbf{AVCL(Ours)}  & RGB+Audio & RGB+Audio & R(2+1)D & \textbf{77.2} & \textbf{67.1}    \\
    \hline
  \end{tabular}
    \vspace{-10pt}
  \label{tab:4-1}
\end{table}

\textbf{The number of models.} From Table \ref{tab:4-2}, we can see that two separated models can better model heterogeneous modalities and perform better than that of using one model.

\textbf{The CGRA module.} It can be observed in Table \ref{tab:4-4} that our AVCL performs better than the AVCL without the CGRA. When the CGRA module is applied in the pool-final layer, the accuracy is the best. This verifies that the CGRA module can effectively align the high-level features of the two modalities.

\textbf{The SelfCL module.} To analyze the contribution of CGRA and SelfCL, we set different values to $\lambda_{cor}$ and $\lambda_{self}$, as shown in Table \ref{tab:4-5}. When setting the weight values of $\lambda_{cor}$ and $\lambda_{self}$ to 0.9 and 0.1, the accuracy is the best. These results validate that SelfCL, CGRA, and AMFM all contribute to the overall framework, while SelfCL contributes more than the CGRA due to the unavailable supervised information.

\textbf{The AMFM module.} It can be observed in Table \ref{tab:4-5} that our AVCL performs better than the AVCL without the AMFM. This verifies that the AMFM module can effectively fuse audio-visual representations.

\vspace{-10pt}
\subsection{Action Recognition}
The action recognition results are shown in Table \ref{tab:4-1}, compared with six state-of-the-art methods, i.e., IIC \cite{28}, MoCo \cite{20}, BYOL \cite{21}, SimCLR \cite{22}, SimSam \cite{23} and Barlow Twins \cite{24}. \textbf{Baseline} is a supervised algorithm to extract features of different modalities through only one R3D network. To facilitate fair comparison, we unify the inputs and outputs as RGB images and audio. Additionally, all comparison methods utilize two models without sharing parameters. Except for our AVCL, all comparison methods merely feed the concatenated features to compute the accuracy.

\begin{figure*}[t]
\centerline{\includegraphics[scale=0.3]{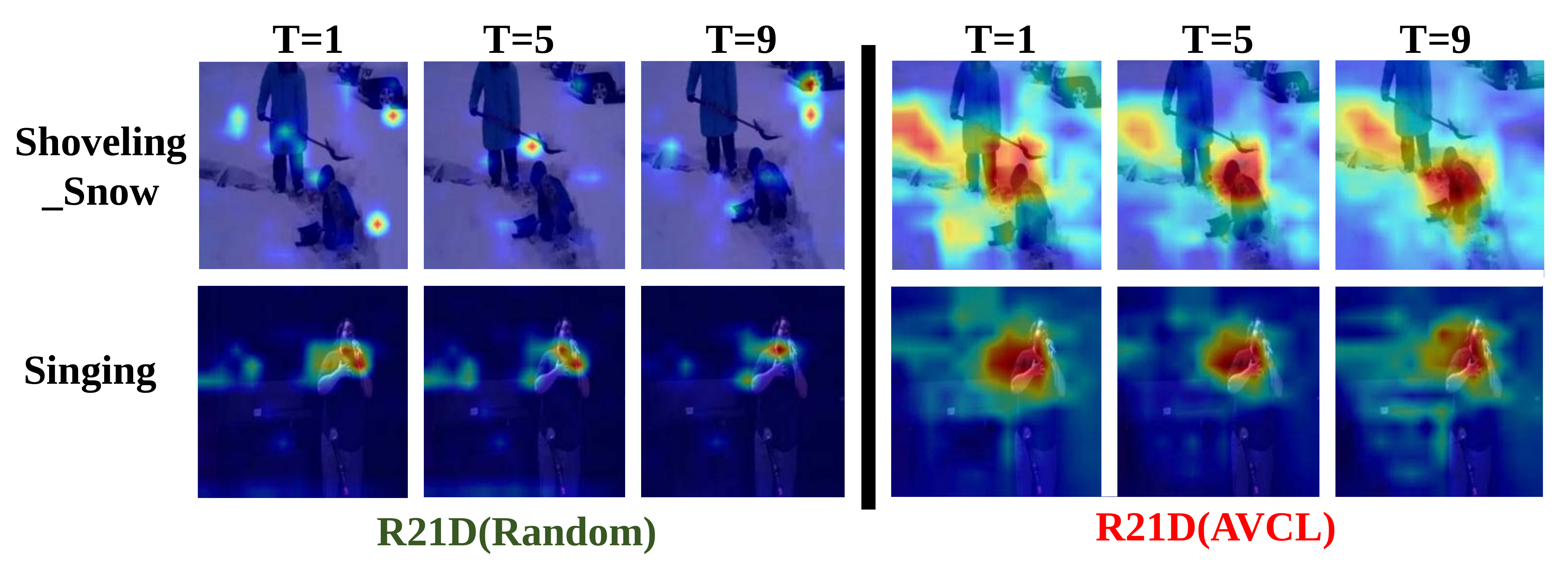}}
  \vspace{-15pt}
\caption{Visualization of a sequence of activation maps on Kinetics-sounds dataset with R(2+1)D as the backbone.}
\vspace{-15pt}
\label{fig:4-1}
\end{figure*}

From Table \ref{tab:4-1}, we can have the following observations: (1) With the same evaluation metric, our AVCL performs favorably against existing approaches under the R3D backbone on both K32 and K100 datasets. (2) Compared with baseline in different modalities, our AVCL achieves significant improvement on both K32 and K100 datasets, which demonstrates the great potential of our AVCL in self-supervised video representation learning and the necessity of using audio modality. (3) After pre-trained with the R3D/C3D/R(2+1)D backbones, we outperform the current contrastive learning methods on both K32 and K100 datasets. This validates the scalability and effectiveness of the AVCL on different backbones.

To have an intuitive understanding of the AVCL, we visualize the spatio-temporal regions using the class activation map, as shown in the Fig. \ref{fig:4-1}. These examples demonstrate strong correlations between highly activated regions and dominant motions in the scene, which validates that our AVCL can effectively capture dominant motions in videos.


 \vspace{-10pt}
\section{Conclusion}
In this paper, we propose a novel framework for self-supervised audio-visual action recognition based on contrastive learning. We first describe attention based multi-modal fusion module and co-correlation guided representation alignment module, and then introduce a self-supervised contrastive learning which is suitable for audio-visual modalities. Extensive experiments have been conducted on Kinetics-Sounds32 and Kinetics-Sounds100 datasets. Our method outperforms the state-of-the-art methods in cross-modal scenarios. The Kinetics-Sounds100 dataset will be made publicly available.

\bibliographystyle{IEEEbib}
\bibliography{refs}

\end{document}